\documentclass[conference]{IEEEtran}
\IEEEoverridecommandlockouts
\usepackage{amsmath,amsfonts}
\usepackage{algorithmic}
\usepackage{array}
\usepackage{textcomp}
\usepackage{stfloats}
\usepackage{url}
\usepackage{verbatim}
\usepackage{graphicx}
\usepackage{cite}
\usepackage{graphicx}
\usepackage{multirow}

\usepackage{multicol}
\usepackage{url}
\usepackage{booktabs}
\usepackage{makecell}
\usepackage{color}
\usepackage{bbding}
\usepackage{algorithm}
\usepackage{amssymb}
\def\BibTeX{{\rm B\kern-.05em{\sc i\kern-.025em b}\kern-.08em
    T\kern-.1667em\lower.7ex\hbox{E}\kern-.125emX}}

\begin{document}
\title{VL-UR: Vision-Language-guided Universal Restoration of Images Degraded by Adverse Weather Conditions\\
\thanks{Dong Yang is the corresponding author (dong.yang@polyu.edu.hk). \IEEEauthorrefmark{1}These authors contributed to the work equllly and should be regarded as co-first authors. }
}

\author{
    \IEEEauthorblockN{1\textsuperscript{st} Ziyan Liu\IEEEauthorrefmark{1}}
    \IEEEauthorblockA{\textit{School of Computer Science,} \\
    \textit{Peking University}\\
    Beijing
    }

\\

    \IEEEauthorblockN{3\textsuperscript{rd} Hushan Yu}
    \IEEEauthorblockA{\textit{School of Computer Science,} \\
    \textit{Peking University}\\
    Beijing}
    
\and
    \IEEEauthorblockN{2\textsuperscript{nd} Yuxu Lu\IEEEauthorrefmark{1}}
    \IEEEauthorblockA{\textit{Department of Logistics and Maritime Studies} \\
    \textit{Hong Kong Polytechnic University}\\
    Hong Kong \\
    }

\\
    \IEEEauthorblockN{4\textsuperscript{th} Dong Yang\IEEEauthorrefmark{2} }
    \IEEEauthorblockA{\textit{Department of Logistics and Maritime Studies} \\
    \textit{Hong Kong Polytechnic University}\\
    Hong Kong \\}
}

\maketitle


\begin{abstract}

    Image restoration is critical for improving the quality of degraded images, which is vital for applications like autonomous driving, security surveillance, and digital content enhancement. However, existing methods are often tailored to specific degradation scenarios, limiting their adaptability to the diverse and complex challenges in real-world environments. Moreover, real-world degradations are typically non-uniform, highlighting the need for adaptive and intelligent solutions. To address these issues, we propose a novel vision-language-guided universal restoration (VL-UR) framework. VL-UR leverages a zero-shot contrastive language-image pre-training (CLIP) model to enhance image restoration by integrating visual and semantic information. A scene classifier is introduced to adapt CLIP, generating high-quality language embeddings aligned with degraded images while predicting degraded types for complex scenarios. Extensive experiments across eleven diverse degradation settings demonstrate VL-UR's state-of-the-art performance, robustness, and adaptability. This positions VL-UR as a transformative solution for modern image restoration challenges in dynamic, real-world environments.
\end{abstract}


\section{Introduction}
    \IEEEPARstart{H}{arsh} imaging environmental conditions, such as haze, low light, rain, and snow, significantly impair the operational effectiveness of various imaging systems, including surveillance, and autonomous vehicles, thereby precipitating a decline in their performance capabilities \cite{gupta2024robust,wang2022low}. The unexpected degradation can significantly compromise the accuracy, reliability, and efficiency of these applications, ultimately hindering their performance and effectiveness. Moreover, these imaging-related issues can result in false alarms, misclassifications, and even system failures, leading to substantial economic losses and safety risks. To address this challenge, researchers are focusing on developing more resilient, efficient, and adaptable multi-scene image restoration (MsIR) methods \cite{li2022all, valanarasu2022transweather, ozdenizci2023restoring, zhu2023learning, guo2024onerestore}. 
    \begin{figure}[t]
        \centering
        \setlength{\abovecaptionskip}{-0.3cm}
        \includegraphics[width=1.00\linewidth]{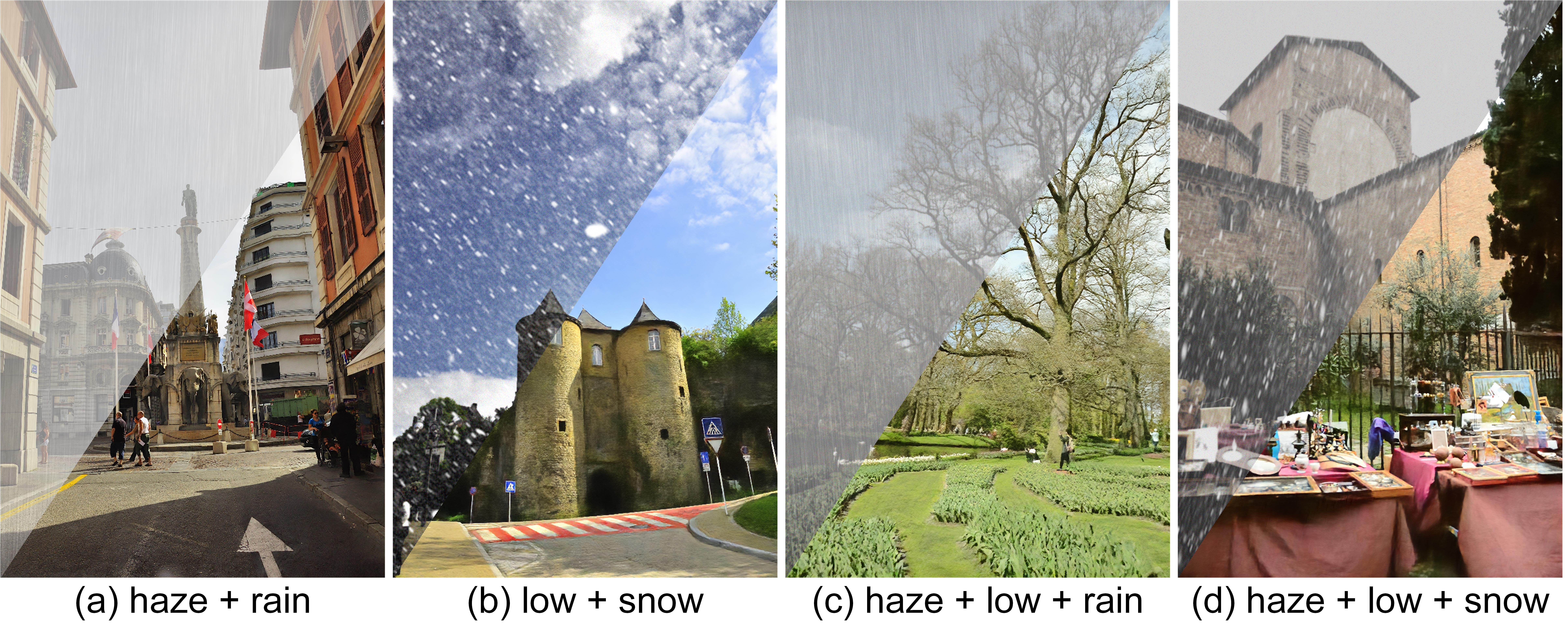}
        \caption{Examples of scene recovery under four different imaging conditions are presented in (a)-(d). The upper triangles in each sub-figure represent the degraded patterns, while the lower triangles display the corresponding restored patterns produced by our method.}
        \label{Figure_flowchart}
        \vspace{-15pt}
    \end{figure}

    MsIR focuses on mitigating image degradation caused by various environmental factors, aiming to restore high-quality images. Traditional methods, such as those relying on handcrafted filters and statistical models \cite{he2010single}, often struggle to capture the complexity and variability of real-world degradation. To address these limitations, learning-based methods have emerged, including end-to-end methods, generative adversarial networks \cite{guo2020joint}, Transformer-based methods \cite{chen2021pre}, diffusion-based models \cite{ozdenizci2023restoring,ye2024learning}, vision-language-guided frameworks \cite{lin2024improving,guo2024onerestore}, and similar methods. For instance, Li et al. \cite{li2022all} introduced an all-in-one framework capable of handling unknown types of image degradation. Chen et al. \cite{chen2021pre} leveraged pre-trained Transformer models to enhance performance across multiple image processing tasks. PromptIR \cite{potlapalli2024promptir} is a prompt-based learning method, enabling a single model to effectively manage various degradation types and levels without requiring prior knowledge. Ye et al. \cite{ye2024learning} developed a diffusion texture prior model that explicitly captures high-quality texture details, incorporating conditional guidance adapters for realistic and high-fidelity image restoration. Similarly, OneRestore \cite{guo2024onerestore} introduced a cross-attention mechanism that adaptively and controllably restores images with complex composite degradations using textual guidance. However, when images are subjected to mixed and complicated degradation, it remains a significant challenge to recover the underlying detail features.

Recent research has explored using CLIP's feature mapping for lower-level tasks\cite{bai2025textir}. For example, CLIP-IQA \cite{wang2023exploring} demonstrated that the extensive visual-language priors embedded in CLIP can be effectively utilized to assess image quality and abstract perception. In image restoration, current methods mainly focus on single scenarios, which can't fully leverage CLIP's generalization ability. In low-light scenarios, CLIP-LIT\cite{liang2023iterative} guides the network by learning prompts through a complex network design. For multi-scenario cases, DA-CLIP\cite{luo2023controlling} duplicates the CLIP image encoder to enable it to simultaneously encode degraded features and image features. Yet, it doesn't integrate different scenarios, unlike real-world situations where multiple phenomena like snow and fog often co-occur.

\begin{figure*}[htbp]
    \centering
    \setlength{\abovecaptionskip}{-0.1cm}  
    \setlength{\belowcaptionskip}{-3cm}  
    \includegraphics[width=0.9\textwidth]{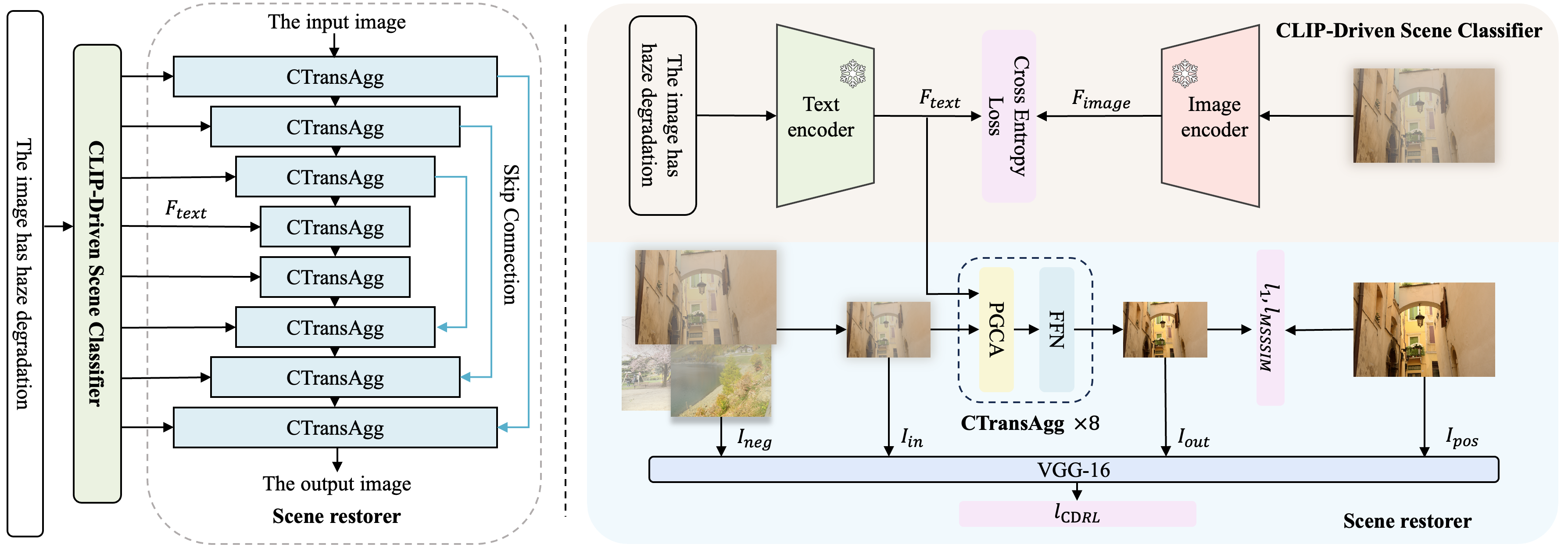}  
    \caption{Overview of our method. VL-UR mainly consists of two parts, a pretrained degraded Scene Classifier (SC)  with the frozen CLIP and a Scene Restorer (SR) network guided by SC. The specific structures of SC and SR are shown on the right side. SC is pretrained and then frozen. Subsequently, during the training phase of SR, SC will extract rich semantic information $\mathbf{F}_{\text{text}}$ from the captions corresponding to the images. Then $\mathbf{F}_{\text{text}}$ will enter the PGCA in CTransAgg for restoration. In the picture, $\mathbf{I}_{\text{in}}$, $\mathbf{I}_{\text{out}}$, $\mathbf{I}_{\text{pos}}$, $\mathbf{I}_{\text{neg}}$ respectively represent the input image, the restored image, the corresponding positive sample and the negative sample of other degradation types. These images will later be used to calculate the loss function.}
    \vspace{-15pt}  
    \label{Figure_flowchart}
\end{figure*}

    In this work, we focus on two key issues: \textbf{\textit{(1) how to effectively leverage the vast semantic information encapsulated in large models to guide image restoration?}} The semantic information encapsulated in large models has a strong ability to generalize across diverse natural environment images. Thus, we leverage the powerful zero-shot capability of CLIP to enhance the deep interaction between semantic and image features. \textbf{\textit{(2) how to enhance processing efficiency to better support downstream tasks?}}  We simplify the training process by moving away from complex prompt engineering. Instead, we focus on fine-tuning CLIP with clear and straightforward prompts. This approach leads to more efficient and faster results which is better suited for downstream tasks. Ultimately, we propose the VL-UR framework, which combines a large-scale pretrained CLIP model with an image restoration network, forming a multi-task framework capable of addressing both degradation-specific and unified image restoration challenges. First, we leverage a zero-shot strategy through CLIP to extract semantic features that are closely aligned with the input image, effectively bridging the gap between corrupted inputs and clean captions. Next, we integrate CLIP's prior knowledge into a self-attention-based restoration network, enhancing the model’s ability to handle complex degradation scenarios. This integration not only improves restoration performance but also enables a unified approach to address diverse degradation types. Finally, the VL-UR model demonstrates robust classification capabilities, accurately identifying eleven distinct types of image degradation. It seamlessly integrates into existing restoration frameworks, delivering visually appealing results across various degradation conditions.
	\begin{itemize}
		\item We present VL-UR, a novel framework that utilizes large-scale pretrained vision-language models as a universal solution for image restoration.
		%
		
		%
        \item By leveraging CLIP's powerful zero-shot capabilities, our model excels in predicting scenes across diverse and complex multi-scene environments. Simultaneously, it matches the most similar semantic features based on the input degradation characteristics and adapts the fixed CLIP text encoder to generate high-quality language embeddings.
		
		\item  Extensive experimental results show that our VL-UR achieves highly competitive performance in all eleven degradation types and can be more applicable to dynamic real-world imaging environments.
	\end{itemize}
\section{Vision Language-guided Universal Restoration Framework}
\label{sec:vl-ur}
\subsection{Overall Network Architecture}
    VL-UR will first retrain the CLIP model within the degraded scene classifier (SC) and then freeze its weights. The SC extracts high-dimensional semantic features and aligns them with the image to support the scene restorer (SR) in enhancing degraded images. As shown in Fig. \ref{Figure_flowchart}, given any type of degraded image $I_{d}$, we utilize the SC to identify its specific degradation type and extract the degradation embedding $\mathbf{F}_{\text{text}}$ from text-encoder $\mathbb{E}_{\text {text}}$ and content embedding $\mathbf{F}_{\text{image}}$ from image-encoder $\mathbb{E}_{\text{image}}$. The embeddings $F_{\text{text}}$ are then integrated into a Transformer-based Encoder-Decoder, which consists of multiple cross-Transformer aggregation blocks at different scales. Finally, to further enhance the restoration performance of the VL-UR, we still suggest the loss function in OneRestore \cite{guo2024onerestore} that combines smooth $\ell_1$ loss, multi-scale structural similarity (MS-SSIM) loss, and composite degradation restoration loss.
\subsection{Zero-shot Degradation Scene Classifier}
\subsubsection{Prompt Ensemble}
%
    For downstream tasks, we focused on efficient processing by avoiding complex prompt training. Multimodal learning theory suggests that linguistic information can enhance a model's semantic understanding, thus improving its performance on specific tasks. In this work, we aim to restore 11 types of degradation, namely: ['\textit{haze}', '\textit{low light}', '\textit{rain}', '\textit{snow}', '\textit{haze + low}', '\textit{haze + rain}', '\textit{haze + snow}', '\textit{low + rain}', '\textit{low + snow}', '\textit{haze + low + rain}', '\textit{haze + low + snow}']. To help our classifier better distinguish these degradation types, we propose a degradation-type prompt template: “\textit{The image has 〈type〉 degradation}”.
    
%
\subsubsection{Inferential Principle}
\begin{figure}[t]
        \centering
        \setlength{\abovecaptionskip}{-0.cm}
        \includegraphics[width=1.00\linewidth]{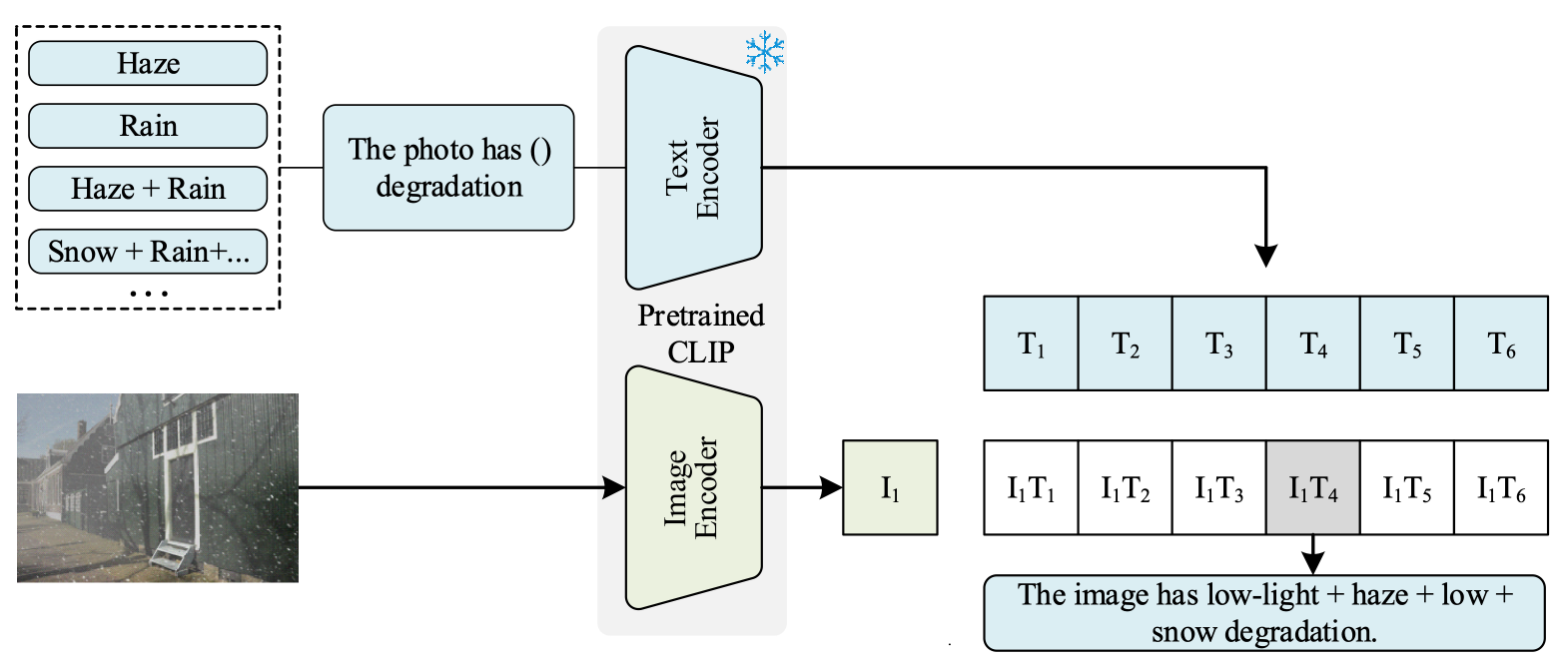}
        \caption{The Composition and Inference Process of Zero-shot Degradation Scene Classifier.}
        \label{Figure_03}
         \vspace{-16pt}  
    \end{figure}

    The core of SC is to obtain degraded type from CLIP, which instructs the network in identifying the optimal neural operation range for regions with different densities, thereby achieving improved aggregation results. Drawing inspiration from Fig. \ref{Figure_03}, we believe that CLIP has the potential to generate degraded types using its powerful zero-shot transfer capability. Specifically, given an unknown degradation type image $I_{d}$ and a prompt set describing imaging conditions $T \in \mathbb{R}^{11}$, both are encoded into the latent space by means of a pretrained CLIP model. Subsequently, the similarity between the image features and each of the text features is calculated. The text corresponding to the maximum similarity value is then designated as the degraded type $T_{d}$. which can be given as 
    \begin{equation}
        \mathbf{F}_{\text{text}}=\mathbb{E}_{\text{text}}(T) \in \mathbb{R}^{11 \times 512},
    \end{equation}
    \begin{equation}        
        \mathbf{F}_{\text{image}}=\mathbb{E}_{\text{image}}(I_d) \in \mathbb{R}^{1 \times 512},
    \end{equation}
    \begin{equation}           
        \begin{aligned}
            \mathbf{T}_{\text{d}} &= \operatorname{Max}(\operatorname{Softmax}(\mathbf{F}_{\text{image}} \cdot \mathbf{F}_{\text{text}}^T)).
        \end{aligned}
    \end{equation}
\subsection{Language-guided Scene Restorer}

\subsubsection{Restoration Branch}
    Given a degraded input $I_d \in \mathbb{R}^{H \times W \times 3}$, the restoration process begins with a $3 \times 3$ convolution to extract shallow embeddings $\mathbf{X}_s \in \mathbb{R}^{H \times W \times C}$, where $H$ and $W$ denote the spatial dimensions and $C$ is the number of channels. The shallow features $\mathbf{X}_s$ are then hierarchically encoded into deep features $\mathbf{X}_l^{e, d} \in \mathbb{R}^{\frac{H}{2^{l-1}} \times \frac{W}{2^{l-1}} \times 2^{l-1} C}$. After encoding the degraded input into latent features $\mathbf{X}_4 \in \mathbb{R}^{\frac{H}{8} \times \frac{W}{8} \times 8 C}$, the decoder progressively recovers high-resolution representations. Finally, a $3 \times 3$ convolution reconstructs the restored image $I_r \in \mathbb{R}^{H \times W \times 3}$. In the restoration branch, we build upon OneRestore \cite{guo2024onerestore} and propose Cross-Transformer Aggregation (CTransAgg), which consists of two components: Prompt Guidance Cross-attention (PGCA), and Feed-Forward Network (FFN). The restoration process can be described as
    \begin{equation}
        \mathcal{F}_{t b}\left(\mathbf{X}_l^e\right)=\mathcal{F}_f\left(\mathcal{F}_p\left(\mathbf{X}_l^e\right)\right),
    \end{equation}
    \begin{equation}
        I_r=\mathcal{C}\left(\mathcal{F}_{t b}^{4,3,2,1}\left(\mathcal{F}_{t b}^{1,2,3,4}\left(I_d\right) \downarrow_{\times 3}, \mathbf{X}_l\right)\uparrow_{\times 3}\right) + I_d,
    \end{equation}
    where $\mathcal{F}_{\mathrm{f}}(\cdot)$ and $\mathcal{F}_{\mathrm{p}}(\cdot)$ represent the FFN, PGCA processes respectively. $\mathcal{C}(\cdot)$ denotes the $3 \times 3$ convolution operation, and $\downarrow$ and $\uparrow$ indicate down-sampling and up-sampling, respectively.
\subsection{Prompt Guidance Cross-attention Module}
%
    We propose a novel PGCA which is designed to assist the network in capturing and preserving structural semantic information. It can be simultaneously guided by the information from both images and texts. Notice that given a degraded image $\mathbf{I}_d$ and a set of non-corresponding degradation types $\mathbf{T}_s$ (including all possible types), we can determine the image’s corresponding degradation type $\mathbf{T}_d$ through a pretrained SC encoder. Since the predictions from the pretrained SC encoder are not always fully accurate and may introduce errors, we adopt a phased strategy. During training, we input the specified text directly into the pretrained SC encoder to obtain the corresponding feature $\mathbf{Y}_T$. During testing, we offer the flexibility to either use the pretrained SC encoder's predictions or provide the specified text manually. 
    
    PGCA takes as input a layer-normalized defect image feature tensor $\mathbf{I} \in \mathbb{R}^{\hat{H} \times \hat{W} \times \hat{C}}$ and the text feature tensor $\mathbf{Y}_T$ that we have just obtained. It generates $\mathbf{Q}_I$, $\mathbf{K}_I$, and $\mathbf{V}_I$ projections using $1 \times 1$ convolutions followed by $3 \times 3$ depthwise convolutions to enrich both pixel-wise and channel-wise context. Additionally, $\mathbf{Q}_T$ is obtained from $\mathbf{Y}_T$ through a linear mapping

    \begin{equation}
    \begin{aligned}
    \operatorname{PGCA}({{Q_I}}, {{Q_T}},{{K_I}}, {{V_I}}) &= {{V_I}} \cdot (\operatorname{Softmax}({{K_I}} \cdot {{Q_I}} / \alpha)\\
    & \quad +\operatorname{Softmax}({{K_I}} \cdot {{Q_T}} / \alpha))
    \end{aligned}
    \end{equation}
    where $\mathbf{\alpha}$ is a learnable scaling parameter, and like conventional multi-head self-attention, the channels in PGCA are divided into 'heads' to learn separate attention maps in parallel.
    \begin{figure}[t]
        \centering
        \setlength{\abovecaptionskip}{0.cm}
        \includegraphics[width=1.0\linewidth]{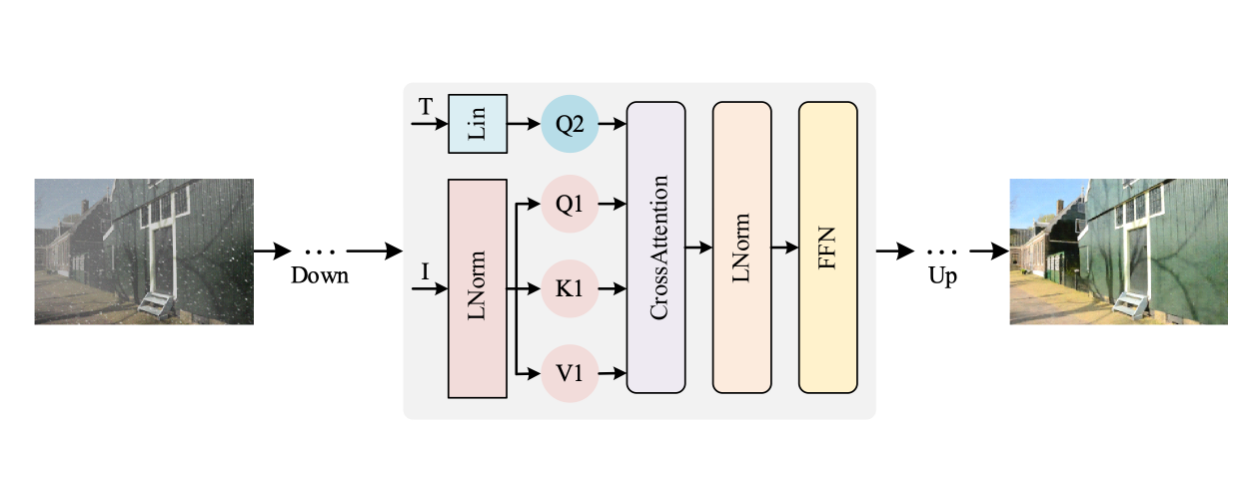}
        \caption{The structure of the CtransAgg module includes the PGCA and FFN, where the PGCA contains the cross attention and LNorm.}
        \label{Figure_04}
        \vspace{-16pt}
    \end{figure}
\subsection{Feed-Forward Network}
    The FeedForward Network (FFN) effectively enhances feature processing by combining channel interaction and spatial context modeling. First, the input features are projected into a higher-dimensional space using a $1 \times 1$ convolution, capturing inter-channel correlations. Next, a depthwise separable $3 \times 3$ convolution models feature dependencies within local spatial regions and splits the result into two parts along the channel dimension: one part is activated by the GELU function, while the other serves as a modulation factor. The two parts are fused through element-wise multiplication, and the final output is projected back to the input dimension via another $1 \times 1$ convolution. The overall process can be expressed as
    \begin{equation}
        \text{FFN}(x) = \text{Conv}_{1 \times 1} ( \text{GELU}\big(x_1\big) \odot x_2),
    \end{equation}
    where $x_1, x_2 = \text{Chunk} (\text{DWConv}_{3 \times 3}( \text{Conv}_{1 \times 1}(x)))$. This structure efficiently fuses channel and spatial information, significantly improving the model's ability to capture both local details and global features in image restoration tasks.
    \setlength{\tabcolsep}{2.60pt}
    \begin{table*}[t]
        \centering
        \caption{Comparison of deraining quantitative results (PSNR / SSIM) on CDD-11 \cite{guo2024onerestore}. The best results are in \textbf{bold}, and the second best are with \underline{underline}.}
        \begin{tabular}{l|ccc|ccccccc}
        \hline
        & \multicolumn{3}{c|}{One-to-One} & \multicolumn{6}{c}{All-in-One} \\
        & MIRNet \cite{zamir2022learning} & Fourmer \cite{zhou2023fourmer} & OKNet \cite{cui2024omni} & AirNet \cite{li2022all} & TransW \cite{valanarasu2022transweather} & Diffusion \cite{ozdenizci2023restoring} & Prompt \cite{potlapalli2024promptir} & WGWS \cite{zhu2023learning}  & OneRe \cite{guo2024onerestore} & VL-UR\\ \hline\hline
        I    & 25.30    / 0.963 & 22.35    / 0.923 & 28.27    / 0.983 & 24.21    / 0.955 & 23.95    / 0.923 & 21.99    / 0.900 & 26.10    / 0.97 & 27.90    / 0.983 & 32.71    / 0.991 & 33.14    / 0.990   \\
        II   & 26.09  / 0.793   & 24.91  / 0.756   & 25.64  / 0.807   & 24.83  / 0.773   & 23.39  / 0.719   & 23.58  / 0.753   & 26.33  / 0.799  & 24.39  / 0.768   & 26.55    / 0.819 & 26.40  / 0.819     \\
        III  & 31.25  / 0.936   & 28.76  / 0.893   & 30.98  / 0.949   & 26.55  / 0.898   & 26.69  / 0.899   & 24.85  / 0.883   & 31.56  / 0.949  & 33.15  / 0.965   & 33.48    / 0.965 & 33.63  / 0.966     \\
        IV   & 31.20  / 0.956   & 28.28  / 0.929   & 32.35  / 0.966   & 26.79  / 0.927   & 25.74  / 0.894   & 24.80  / 0.887   & 31.53  / 0.964  & 34.43  / 0.975   & 34.50    / 0.976 & 34.71  / 0.976     \\
        V    & 23.35  / 0.763   & 21.75  / 0.731   & 24.08  / 0.802   & 23.23  / 0.775   & 22.24  / 0.715   & 21.83  / 0.746   & 24.49  / 0.784  & 24.27  / 0.794   & 26.15    / 0.822 & 26.15  / 0.822     \\
        VI   & 24.27  / 0.906   & 21.68  / 0.847   & 25.83  / 0.936   & 22.21  / 0.875   & 23.11  / 0.876   & 21.25  / 0.865   & 24.54  / 0.927  & 27.23  / 0.955  & 30.27    / 0.960 & 30.39  / 0.962     \\
        VII  & 23.83  / 0.922   & 21.20  / 0.880   & 27.14  / 0.953   & 23.29  / 0.910   & 22.34  / 0.872   & 21.99  / 0.867   & 23.70  / 0.930  & 27.65  / 0.962   & 30.46    / 0.968 & 30.26  / 0.968     \\
        VIII & 24.75  / 0.755   & 23.48  / 0.692   & 24.54  / 0.768   & 22.82  / 0.709   & 22.62  / 0.691   & 22.69  / 0.722   & 25.05  / 0.767  & 25.06  / 0.766   & 25.83    / 0.796 & 25.75  / 0.797     \\
        IX   & 24.33  / 0.744   & 23.28  / 0.704   & 24.20  / 0.763   & 23.29  / 0.724   & 21.80  / 0.662   & 22.12  / 0.702   & 24.51  / 0.759  & 24.60  / 0.760   & 25.56    / 0.791 & 25.40  / 0.790     \\
        X    & 22.43  / 0.717   & 21.21  / 0.677   & 23.13  / 0.761   & 21.80  / 0.707   & 21.55  / 0.674   & 21.23  / 0.709   & 23.74  / 0.748  & 23.90  / 0.766   & 25.18    / 0.789 & 25.25  / 0.790     \\
        XI   & 22.27  / 0.721   & 20.91  / 0.684   & 23.48  / 0.762   & 22.24  / 0.726   & 21.01  / 0.656   & 21.04  / 0.694   & 23.33  / 0.745  & 23.97  / 0.767   & 25.18    / 0.789 & 25.24  / 0.791    \\  \hline
        Avg.             & 25.37 / 0.834 & 23.44 / 0.792 & 26.33 / 0.859 & 23.75 / 0.816 & 23.13 / 0.780 & 22.49 / 0.793 & 25.90 / 0.849 & 26.96 / 0.860  & 28.72    / 0.879 & 28.76 / 0.879 \\  \hline        
        \end{tabular}\label{table:cdd}
        \footnotesize{\noindent I: haze, II: low light, III: rain, IV: snow, V: haze+low, VI: haze+rain, VII: haze+snow, VIII: low+rain, IX: low+snow, X: haze+low+rain, XI: haze+low+snow}
        \vspace{-0.5cm} 
    \end{table*}
\subsection{Loss Function}
    In this work, we suggest a hybrid loss function that integrates Smooth ${L}_1$ loss, Multi-Scale Structural Similarity (MS-SSIM) loss, and  Composite Degradation Restoration Loss (CDRL), aiming to achieve multi-level optimization at the pixel, structural, and feature levels. The definitions and function of each component loss function are described as follows.
\subsubsection{Smooth ${L}_1$ Loss}
    Smooth ${L}_1$ Loss is used to measure the pixel-wise differences between the restored image $I_r$ and the ground truth image $I_g$, which can be defined as
    \begin{equation}
        \mathcal{L}_{\text{SmoothL1}} = \frac{1}{N} \sum_{k=1}^N \operatorname{SmoothL1} \left(I_r(k) - I_g(k) \right),
    \end{equation}
    where $N$ is the total number of pixels in the image, and $I_r(k)$ and $I_g(k)$ represent the pixel values of the restored and ground truth images at the $k$-th position, respectively. Compared to traditional ${L}_1$ or ${L}_2$ loss, Smooth ${L}_1$ Loss is more robust. It behaves as a quadratic loss for small errors, providing finer gradient information, and transitions to a linear loss for larger errors, thereby preventing gradient explosion during training.
\subsubsection{MS-SSIM Loss}
    The MS-SSIM evaluates the structural similarity between the restored image and the ground truth image across multiple scales. The MS-SSIM loss is defined as
    \begin{equation}
        \mathcal{L}_{\text{MSSSIM}} = 1 - \prod_{i=1}^{L} \left( \text{SSIM}_i \right)^{w_i},
    \end{equation}
    where $L$ represents the total number of scales, $\text{SSIM}_i$ is the structural similarity at the $i$-th scale, and $w_i$ is the weight for the corresponding scale, satisfying $\sum w_i = 1$. The structural similarity $\text{SSIM}_i$ at each scale is calculated using the local luminance, contrast, and structural information of the images. In our implementation, the weights $w_i$ are predefined as $[0.0448, 0.2856, 0.3001, 0.2363, 0.1333]$. By minimizing $1 - \text{MS-SSIM}$, the model can better preserve the multi-scale structural information of the restored image.
\subsubsection{CDRL Loss}
    The CDRL Loss enhances the feature separation of the restored image by constraining its high latitude feature distance from positive and negative samples. We use VGG-16 to extract the features of high latitude. The mathematical definition of CDRL Loss is
    \begin{equation}
        \mathcal{L}_{\text{CDRL}} = \sum_{i=1}^K \frac{\operatorname{L}_1\left(\mathbf{v}_{\text{rec}, i}, \mathbf{v}_{\text{pos}, i}\right)}{\lambda_1 \operatorname{L}_1\left(\mathbf{v}_{\text{rec}, i}, \mathbf{v}_{\text{input}, i}\right) + \sum_{j=1}^N \lambda_2 \operatorname{L}_1\left(\mathbf{v}_{\text{rec}, i}, \mathbf{v}_{\text{neg}, j}\right)},
    \end{equation}
    where $\mathbf{v}_{\text{rec}, i}$ is the feature extracted from the restored image after passing through the $i$-th layer of VGG-16, $\mathbf{v}_{\text{pos}, i}$ is the feature of the positive sample (ground truth image) at the same layer, \(\mathbf{v}_{\text{input}, i}\) is the feature of the input image, and $\mathbf{v}_{\text{neg}, j}$ is the feature of the $j$-th negative sample. $N$ is the number of negative samples, while $\lambda_1$ and $\lambda_2$ are balancing coefficients that control the influence of positive and negative samples on the loss. By minimizing the CDRL Loss, the model can learn to reduce the distance between the restored image and the positive sample while increasing the distance from the negative samples, thus improving the feature separation of the restored image.
\subsubsection{Total Loss}
    To integrate the above loss functions, we define the total loss function $\mathcal{L}_{\text{total}}$, which can be given as
    \begin{equation}
        \mathcal{L}_{\text{total}} = \gamma_1 \mathcal{L}_{\text{SmoothL1}} + \gamma_2 \mathcal{L}_{\text{MSSSIM}} + \gamma_3 \mathcal{L}_{\text{CDRL}},
    \end{equation}
    where $\gamma_1$, $\gamma_2$, and $\gamma_3$ are the weights for the Smooth ${L}_1$ Loss, MS-SSIM Loss, and CDRL Loss, respectively, based on experimental results, the optimal weights are set as $\gamma_1 = 0.6$, $\gamma_2 = 0.3$, and $\gamma_3 = 0.1$. By minimizing the total loss function $\mathcal{L}_{\text{total}}$, the model achieves comprehensive optimization at the pixel, structural, and feature levels, significantly enhancing the quality and performance of the restored images. More details about loss function can be found in \cite{guo2024onerestore}.
\section{Experiments and Discussion}
\label{sec:experiments}
    In this section, we perform extensive experiments to showcase the exceptional low-visibility restoration capabilities of VL-UR. 
    
%

\subsection{Dataset and Implementation Details}
\subsubsection{Competitive methods}
    To evaluate the effectiveness of our method, we conduct a comparative analysis between VL-UR and various state-of-the-art methods, including MIRNet \cite{zamir2022learning}, Fourmer \cite{zhou2023fourmer}, OKNet \cite{cui2024omni}, AirNet \cite{li2022all}, TransW \cite{valanarasu2022transweather}, Diffusion \cite{ozdenizci2023restoring}, Prompt \cite{potlapalli2024promptir}, WGWS \cite{zhu2023learning}, and OneRestore -- OneRe \cite{guo2024onerestore}. Furthermore, to ensure the integrity and objectivity of the experiments, all compared methods used in the study are obtained exclusively from the author's available source code.

\subsubsection{Datasets and Evaluation Metrics}
	%
    The training and testing dataset comprises a composite degradation dataset (CDD-11) \cite{guo2024onerestore}. To quantitatively evaluate the recovery and enhancement performance of different methods, we have chosen a set of evaluation metrics. Two metrics include reference evaluation metrics (i.e., peak signal-to-noise ratio (PSNR), structural similarity (SSIM). A higher value of PSNR and SSIM indicates superior performance in image recovery. Furthermore, all evaluation metrics values included in the studies are derived based on the RGB channels.
    \begin{figure*}[t]
        \centering
        \setlength{\abovecaptionskip}{0.cm}
        \includegraphics[width=1.00\linewidth]{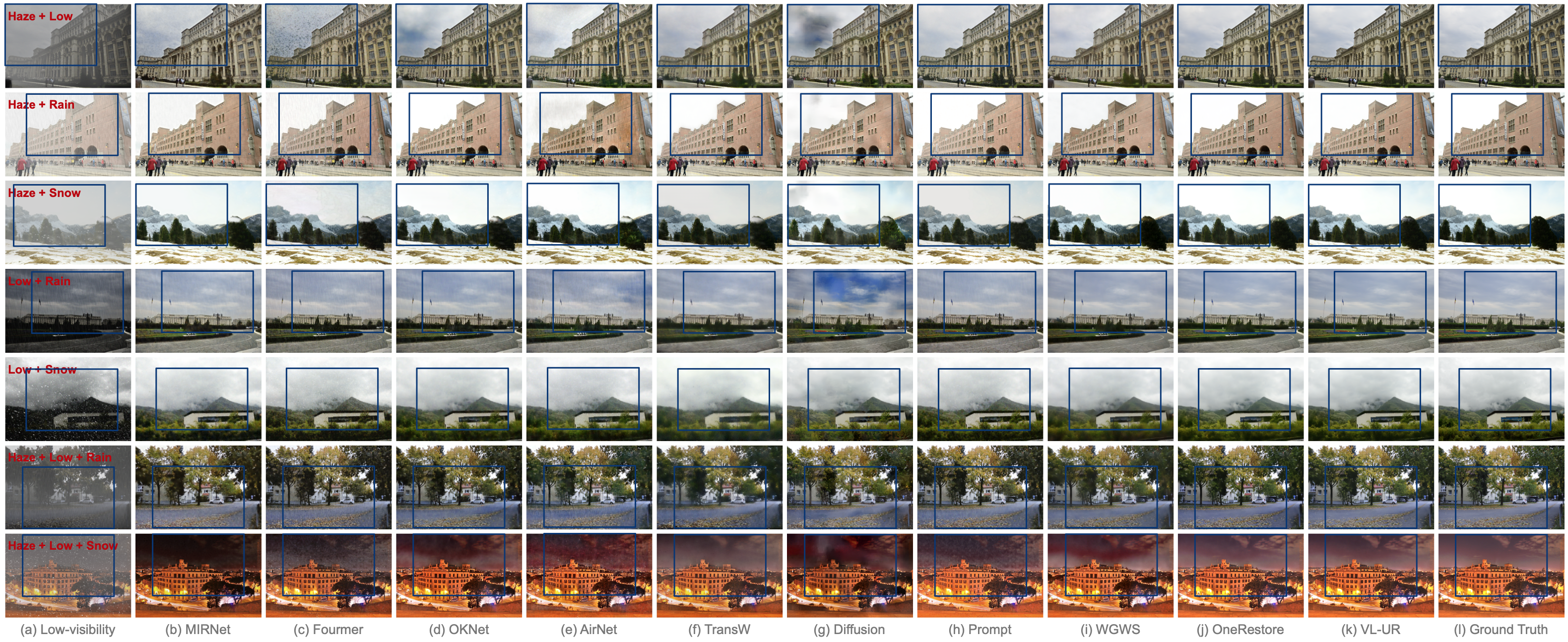}
        \caption{Visual comparisons of multi-scene restoration from CDD-11 \cite{guo2024onerestore}. (a) Low-visibility, restored images, generated by (b) MIRNet \cite{zamir2022learning}, (c) Fourmer \cite{zhou2023fourmer}, (d) OKNet \cite{cui2024omni}, (e) AirNet \cite{li2022all}, (f) TransW \cite{valanarasu2022transweather}, (g) Diffusion \cite{ozdenizci2023restoring}, (h) Prompt \cite{potlapalli2024promptir}, (i) WGWS \cite{zhu2023learning}, (j) OneRestore \cite{guo2024onerestore}, (k) our VL-UR, and (l) Ground Truth, respectively. Zooming in on the image facilitates more precise visual comparison.}
        \label{Figure_Visual}
    \end{figure*}
\subsubsection{Experiment Platform}
    The VL-UR is trained for 200 epochs with CDD-11 train dataset. The ADAM optimizer is responsible for updating the network parameters. The initial learning rate of VL-UR is set to $1 \times 10^{-3}$ and decreases gradually as training progresses. The VL-UR model was trained and evaluated within the Python 3.7 environment using the PyTorch software package with an Intel(R) Core(TM) i9-12900HX CPU @2.30GHz and Nvidia GeForce RTX 4090.

\subsection{CLIP Output Analysis}
    We test the predictive results of SC under each degradation type, achieving an accuracy of up to 94\%. Notably, for more complex scenarios such as "low light + haze + snow," the prediction accuracy was even higher than for single-category scenarios like "haze." This phenomenon can be attributed to the vast amount of image and text information encapsulated within CLIP, which closely mirrors natural environmental conditions. In real-world scenarios, hazy weather often coexists with low light, and heavy snow frequently creates a misty, haze-like visual effect. When text features are extracted by CLIP, single-scene descriptions tend to exhibit a degree of convergence, potentially limiting distinctiveness. To address this, we specifically assigned carefully designed textual prompts to our SC during the training phase. This method leverages the richness of linguistic information encoded in CLIP while avoiding the introduction of erroneous features. As a result, our method achieves superior performance in multi-scene image restoration tasks.
\subsection{Quantitative Analysis and Comparison}
    We evaluate the performance of our VL-UR on mixed degraded images from the CDD-11 dataset \cite{guo2024onerestore} using a set of objective evaluation metrics. As shown in Table \ref{table:cdd}, VL-UR consistently achieves superior results, outperforming all competing methods across every metric. Restoring scenes affected by unknown degradation represents a particularly challenging task. A visual comparison in Fig. \ref{Figure_Visual} highlights the limitations of existing approaches in addressing these compound degradations. In contrast, VL-UR exhibits remarkable robustness, delivering significantly improved results for images with complex mixed degradations.
\subsection{Ablation Study}
\label{ss:as}
\subsubsection{Effectiveness of Network Modules}
    To validate the effectiveness of the PGCA module in capturing key scene features and utilizing degraded scene descriptors, we compared CLIP-guided scene descriptions with the traditional cross-attention method. Results indicate that PGCA enhances PSNR by approximately 1.0 and SSIM by around 0.01, demonstrating its ability to extract critical features and leverage degraded scene descriptors to enhance overall model performance.

    \begin{figure}[t]
        \centering
        \setlength{\abovecaptionskip}{0.cm}
        \includegraphics[width=1.0\linewidth]{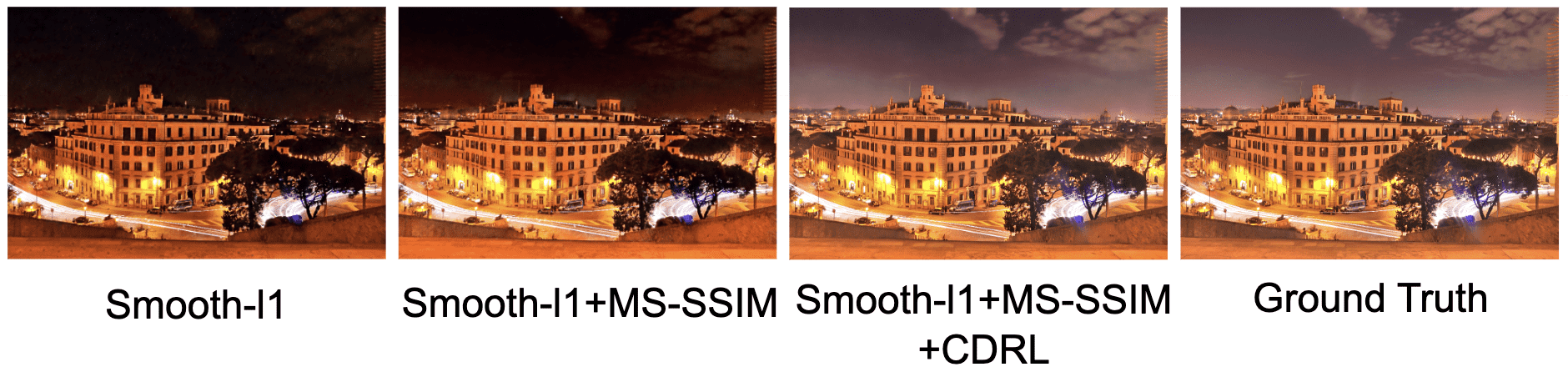}
        \caption{Visualization of different loss functions}
       \vspace{-16pt}
    \end{figure}
    
\subsubsection{Effectiveness of Loss Functions}
    Table \ref{table:loss} presents a performance comparison of different loss functions. The inclusion of the MS-SSIM Loss contributes to a slight increase in SSIM, accompanied by a marginal improvement in PSNR. Conversely, the introduction of CDRL Loss results in a significant boost in PSNR while maintaining a modest enhancement in SSIM. Our experiments indicate that the relative weighting of different loss functions substantially influences image restoration outcomes. A higher proportion of the Smooth L1 Loss ensures that the restored image closely approximates the ground truth, while a smaller proportion of CDRL Loss effectively separates distinct types of degradation features. This balance enables improved performance across varying scene complexities.

    \setlength{\tabcolsep}{10.00pt}
    \begin{table}[t]
        \centering
        \caption{Ablation study for different loss functions.}
        \begin{tabular}{ccc|c}
        \hline
        Smooth $\ell_1$ & MS-SSIM & CDRL     & VL-UR          \\ \hline\hline
        \CheckmarkBold    &         &      & 27.96 / 0.8613         \\ 
        \CheckmarkBold    & \CheckmarkBold     &       & 28.08 / 0.8749 \\ \hline
        \CheckmarkBold    & \CheckmarkBold     & \CheckmarkBold  & 28.75 / 0.8793          \\ \hline
        \end{tabular}\label{table:loss}
    \end{table}
\subsection{Complexity and Running Time Comparison}
    We evaluated the running speed of VL-UR on images with resolutions of 1080p ($1920 \times 1080$) and 2K ($2560 \times 1440$) using a Titan RTX GPU with 24 GB of memory. The results indicate that our method processes 1080p images in approximately 10 milliseconds and 2K images in under 20 milliseconds. These findings highlight the efficiency of our method, demonstrating its ability to meet real-time application requirements while scaling effectively to higher resolutions.
\section{Conclusion}
\label{sec:conclusions}
%

    This work introduces VL-UR, a novel multi-task framework for image restoration that leverages CLIP’s prior knowledge and an image restoration network. By integrating a Scene Classifier, self-attention mechanism, VL-UR accurately classifies eleven degradation types and achieves state-of-the-art performance in both specific and unified restoration tasks. The framework enhances stability, produces visually appealing results, and holds strong potential for downstream applications, making a significant contribution to image restoration

\bibliographystyle{IEEEtran}
\footnotesize
\bibliography{icme.bib}

\end{document}